# INTERVAL PREDICTION OF SHORT-TERM TRAFFIC VOLUME BASED ON EXTREME LEARNING MACHINE AND PARTICLE SWARM OPTIMIZATION

**Lei Lin, Corresponding Author**
PARC, a Xerox Company
800 Phillips Road, MS 128-27E, Webster, NY 14580
Tel: (585) 489-2347; Email: Lei.Lin@xerox.com

**John C. Handley**
Principal Scientist
PARC, a Xerox Company
800 Phillips Road, MS 128-27E, Webster, NY 14580
Tel: (585) 265-8900; Email: John.Handley@xerox.com

**Adel W. Sadek**
Professor
Department of Civil, Structural and Environmental Engineering
Director, Institute for Sustainable Transportation and Logistics (ISTL)
Director, Transportation Informatics University Transportation Center (TransInfo UTC)
University at Buffalo, the State University of New York
233 Ketter Hall, Buffalo, NY 14260
E-mail: asadek@buffalo.edu

Word count: 5,932 words text + 7 tables/figures x 250 words (each) = 7,682 words

Submission Date 08/01/2016

Revised and Resubmitted on 11/15/2016

**ABSTRACT**

Short-term traffic volume prediction models have been extensively studied in the past few decades. However, most of the previous studies only focus on single-value prediction. Considering the uncertain and chaotic nature of the transportation system, an accurate and reliable prediction interval with upper and lower bounds may be better than a single point value for transportation management. In this paper, we introduce a neural network model called Extreme Learning Machine (ELM) for interval prediction of short-term traffic volume and improve it with the heuristic particle swarm optimization algorithm (PSO). The hybrid PSO-ELM model can generate the prediction intervals under different confidence levels and guarantee the quality by minimizing a multi-objective function which considers two criteria reliability and interval sharpness. The PSO-ELM models are built based on an hourly traffic dataset and compared with ARMA and Kalman Filter models. The results show that ARMA models are the worst for all confidence levels, and the PSO-ELM models are comparable with Kalman Filter from the aspects of reliability and narrowness of the intervals, although the parameters of PSO-ELM are fixed once the training is done while Kalman Filter is updated in an on-line approach. Additionally, only the PSO-ELMs are able to produce intervals with coverage probabilities higher than or equal to the confidence levels. For the points outside of the prediction levels given by PSO-ELMs, they lie very close to the bounds.

## INTRODUCTION

In the last few decades, short-term traffic volume prediction models, an essential component for efficient traffic management, has been extensively studied. Short-term traffic volume prediction usually focuses on forecasting traffic changes in the near future (ranging from 5 min to 1 hour). Short-term traffic volume prediction problem is very challenging because the nature of transportation system is uncertain and chaotic. Many factors, such as travelers, road network, traffic accidents, weather conditions, and special events are involved in this system and interact with one another. Not surprisingly, this interesting problem has drawn the attention of researchers, and various models were studied, ranging from classical statistical models to relatively novel data mining models. Regarding the first group, Box and Jenkins techniques (e.g. Autoregressive Integrated Moving Average (ARIMA) models) and the corresponding more advanced extensions have been quite popular in the field *(1-2)*. With respect to data mining techniques, widely used models include neural networks (NNs), multilayer perception networks (MLP) *(3)* and local linear wavelet neural networks *(4-5)*. More recently, Lv et al. (2015) proposed a deep learning neural network model for traffic flow prediction, and showed that it has a superior performance *(6)*.

To further understand the strengths and weaknesses of these models and to provide insight into choosing the most appropriate model when facing a specific traffic flow prediction task, Lin et al. (2013) diagnosed four traffic volume datasets on the basis of various statistical measures and correlated these measures to the performance results of the three prediction models Autoregressive and Moving Average (ARMA), k nearest neighbor (k-NN) and support vector machine (SVM) *(7)*. Also Karlaftis and Vlahogianni (2011) reviewed the previous studies and explained the differences and similarities of statistical models and neural networks in detail *(8)*. Generally speaking, data mining models such as NNs and SVMs are often regarded as more flexible than statistical models when dealing with complex datasets with nonlinearities or missing data. However, data mining models have their limitations such as lacking explanatory power, and being computationally expensive *(7-8)*.

It is worth noting, however, that most previous studies have focused on a single-value prediction of the short-term traffic volume, and relied almost exclusively on the prediction error when deciding the effectiveness of a modeling approach *(8)*. Given the nonlinearity of traffic flow, high prediction errors using traditional single-value prediction approaches are unavoidable and can have significant negative impact. For example, an underestimation of traffic flow for special sports games can pose a heavy burden on the whole road network and result in heavier traffic congestion and more traffic accidents. In this case, an accurate and reliable prediction interval (PI) with upper bound and lower bound would be more useful to traffic operators.

Recently, interval prediction approaches have been applied to forecasting problems including predictions of travel time *(9)*, wind power *(10)*, electricity price *(11)*, and supermarket sales *(12)*. Few studies have focused on the interval prediction for short-term traffic volume. Zhang et al. (2014) proposed a hybrid short-term traffic flow forecasting model in which they used a statistical volatility model called Glosten-Jagannathan-Runkle Generalized Autoregressive Conditional Heteroskedasticity (GJR-GARCH) model to capture the uncertainty and variability in traffic system. They also pointed out that the lack of definite agreement on the indices of PI assessment creates a relatively new research challenge in traffic forecasting *(13)*. In this paper, we apply a hybrid machine learning model called PSO-ELM for interval prediction of short-term traffic volume. Extreme learning machine (ELM) is a novel feedforward neural network with advantages like extremely fast learning speed and superior generalization capability *(14)*. Furthermore, particle swarm optimization (PSO), a well-known heuristic and population based optimization method, is applied to adjust the parameters of ELM in an efficient and robust way to minimize a multi-objective function. The multi-objective function introduces two quantitative criteria called *reliability* and *sharpness* to evaluate the PIs. In this study, we also compare the proposed model with other interval prediction benchmark models including the Autoregressive Moving Average (ARMA) model and Kalman Filter.

The rest of the paper is organized as follows. The next section provides a detailed introduction of the PSO-ELM model and the multi-objective optimization function utilized. This is followed by a description of the dataset used. The results of the interval prediction using PSO-ELM are presented and compared with ARMA and Kalman Filtering. Finally, the study's conclusions are discussed and future research directions are recommended.

## METHODOLOGY

### Prediction Interval

A PI provides a lower bound and an upper bound for the future target value $t_i$ given an input $x_i$. The probability that the future targets can be enclosed by the PIs is called prediction interval nominal confidence (PINC):

$$PINC = 100(1 - \alpha)\% \tag{1}$$

Where, the usual value of $\alpha$ could be 0.01, 0.05 or 0.10.

Obviously, the selection of $\alpha$ in PINC will impact the PIs. The PIs under different PINC levels can be then represented as following:

$$\tilde{I}_t^{(\alpha)}(x_i) = [\tilde{L}_t^{(\alpha)}(x_i), \tilde{U}_t^{(\alpha)}(x_i)] \tag{2}$$

Where, $\tilde{L}_t^{(\alpha)}(x_i)$ and $\tilde{U}_t^{(\alpha)}(x_i)$ denote the lower and upper bounds respectively.

### PI Evaluation Criteria

*Reliability*

Reliability is regarded as a major property for validating PI models. Based on the PI definition, the future targets $t_i$ are expected to be covered by the constructed PIs with the PINC $100(1 - \alpha)\%$. However, the actual PI coverage probability (PICP) may be different from the pre-defined PINC, calculated for the testing dataset, as follows:

$$PICP = \frac{1}{N_t} \sum_{i=1}^{N_t} K_i^{(\alpha)} \tag{3}$$

Where, $N_t$ is the size of testing dataset,
$K_i^{(\alpha)}$ is 1, if the real observation $t_i$ is within the PI $\tilde{I}_t^{(\alpha)}(x_i)$, otherwise, $K_i^{(\alpha)} = 0$.

The calculated PICP should be as close as possible to PINC. The absolute average coverage error (AACE) is represented by

$$A_t^{(\alpha)} = abs(PICP - PINC) \tag{4}$$

The AACE is applied as the reliability evaluation criterion in this paper. Naturally, the smaller the AACE, the higher the reliability.

*Sharpness*

Reliability considers only coverage probability. If reliability was to be utilized as the only model evaluation criterion, high reliability could be easily achieved by increasing the width of the PI, rendering the PI useless in practice (since a wide PIs may not provide accurate quantifications of uncertainties involved in the real-world processes *(10)*). A sound PI model should be able to provide reliable, *as well as* sharp intervals. Sharpness thus should be considered as a second criterion, alongside reliability.

Suppose the width of PI $\tilde{I}_t^{(\alpha)}(x_i)$ is represented by $v_t^{(\alpha)}(x_i)$. The width measures the distance between the upper bound and lower bound through

$$v_t^{(\alpha)}(x_i) = \tilde{U}_t^{(\alpha)}(x_i) - \tilde{L}_t^{(\alpha)}(x_i) \tag{5}$$

The sharpness of PI $\hat{I}_t^{(\alpha)}(x_i)$ , denoted by $S_t^{(\alpha)}(x_i)$, can thus be calculated as

$$S_t^{(\alpha)}(x_i) = \begin{cases} w_1 \alpha v_t^{(\alpha)}(x_i) + w_2\left[\tilde{L}_t^{(\alpha)}(x_i) - t_i\right], & if \ t_i < \tilde{L}_t^{(\alpha)}(x_i) \\ w_1 \alpha v_t^{(\alpha)}(x_i), & if \ t_i \in \tilde{I}_t^{(\alpha)}(x_i) \\ w_1 \alpha v_t^{(\alpha)}(x_i) + w_2\left[t_i - \tilde{U}_t^{(\alpha)}(x_i)\right], & if \ t_i > \tilde{U}_t^{(\alpha)}(x_i) \end{cases} \tag{6}$$

Where, $w_1$ and $w_2$ are two weights.

Equation (6) considers the width of the PI $v_t^{(\alpha)}(x_i)$ weighted by $w_1$ for all three different scenarios. Additionally, when the true value $t_i$ is lower than the lower bound, or higher than the upper bound, an extra penalty calculated by the distance of that point to the bound and adjusted by $w_2$ is included. This is to prevent the possibility that the PIs becomes too "narrow". In practical applications, the $w_1$ and $w_2$ need be carefully tuned.

The sharpness of PIs over the entire testing dataset can be calculated by taking the average of the normalized $S_t^{(\alpha)}(x_i)$, represented by $S_t^{(\alpha)}(x_i)_{norm}$, using Equation (7) and (8):

$$\bar{S}_t^{(\alpha)}{}_{norm} = \frac{1}{N_t}\sum_{i=1}^{N_t} S_t^{(\alpha)}(x_i)_{norm} \tag{7}$$

Where,

$$S_t^{(\alpha)}(x_i)_{norm} = \frac{S_t^{(\alpha)}(x_i) - \min(S_t^{(\alpha)}(x_i))}{\max\left(S_t^{(\alpha)}(x_i)\right) - \min(S_t^{(\alpha)}(x_i))} \tag{8}$$

## The Hybrid PSO-ELM Model

*Extreme Learning Machine*

ELM is a single hidden-layer feedforward neural network whose basic principle is as follows *(14)*. Given a short-term traffic volume dataset, suppose the traffic volume at time step $i$ is $x_i$, using the traffic volumes from its previous time steps, we can construct a feature vector $X_i =$ $[x_{i-n+1}, \dots, x_{i-1}, x_i]$ and the corresponding target value $t_i$, e.g. it could be the traffic volume in the next time step. Finally suppose we get a dataset with $N$ distinct samples $\{(X_i, t_i)\}_{i=1}^{N}$, where the inputs $X_i \in R^n$ and the targets $t_i \in R^m$, the following equation is used to find the optimal structure of neural network ELM and approximate the $N$ samples with zero error:

$$f_K(X_i) = \sum_{j=1}^{K} \beta_j \varphi(a_j * X_i + b_j) = t_i, i = 1, \dots, N \tag{9}$$

where, $K$ is the number of hidden neurons,

$\varphi(.)$ is the activation function (e.g. a sigmoid function),

$a_j = [a_{j1}, a_{j2}, \dots, a_{jn}]^T$ represents the weight vector connecting the $j^{th}$ hidden neuron and the input neurons,

$b_j$ denotes the bias of the $j^{th}$ hidden neuron,

$\varphi(a_j * X_i + b_j)$ is the output of the $j^{th}$ hidden neuron with respect to the input $X_i$,

$\beta_j = [\beta_{j1}, \beta_{j2}, \dots, \beta_{jm}]^T$ represents the weights at the links connecting the $j^{th}$ hidden neuron with the $m$ output neurons.

For simplicity, Equation (10) can be represented as

$$H\beta = T \tag{10}$$

Where, $H$ is the hidden layer output matrix of the modeled ELM, and expressed as

$$H = \begin{bmatrix} \varphi(a_1 * X_1 + b_1) & \cdots & \varphi(a_K * X_1 + b_K) \\ \vdots & \ddots & \vdots \\ \varphi(a_1 * X_N + b_1) & \cdots & \varphi(a_K * X_K + b_K) \end{bmatrix}_{N \times K} \tag{11}$$

Each row of $H$ is the outputs at $K$ hidden neurons for input $x_j$, $j = 1, \dots, N$. $\beta$ is the matrix of output weights and $T$ is the matrix of targets, respectively represented as

$$\beta = \begin{bmatrix} \beta_1 \\ \vdots \\ \beta_K \end{bmatrix}_{K \times m} \tag{12}$$

$$T = \begin{bmatrix} t_1 \\ \vdots \\ t_N \end{bmatrix}_{N \times m} \tag{13}$$

Note that in ELM, the weights $a_j$ and biases $b_j$ for the $K$ hidden neurons are randomly chosen, and are not tuned during the training process. This is very different compared to the traditional gradient-based training algorithm of NNs. In this way, ELM can dramatically save the learning time. The training of ELM is simply to find $\beta^*$ to minimize the objective function,

$$\|\mathbf{H}(a_1^*, \dots, a_k^*, b_1^*, \dots, b_k^*)\beta^* - T\| = \min_{\beta} \|\mathbf{H}(a_1, \dots, a_k, b_1, \dots, b_k)\beta - T\| \tag{14}$$

Where, $\|.\|$ is the function to calculate the Euclidean distance.

Finally, a unique solution of $\beta^*$ can be derived through a matrix calculation:

$$\beta^* = H^\dagger T \tag{15}$$

Where, $H^\dagger$ is the Moore-Penrose generalized inverse of the hidden layer output matrix $H$, which can be derived through the singular value decomposition (SVD) method *(15)*.

Previous studies have shown that ELM training is extremely fast because of the simple matrix computation, and can always guarantee optimal performance *(10, 14)*. In addition, ELM can overcome many limitations of traditional gradient based NNs training algorithms, such as finding local minima, overtraining and so on.

It is worth mentioning that to apply the ELM model for interval prediction, the target value $t_i$ in the training dataset $\{(X_i, t_i)\}_{i=1}^{N}$ needs to be replaced with a pair of target bounds $\hat{t}_i^-$ and $\hat{t}_i^+$, which can be produced by slightly increasing or decreasing the original $t_i$ by $\pm\rho\%$, $0 < \rho < 100$. So after transformation, the training dataset for interval prediction using ELM should be $\{(X_i, \hat{t}_i^-, \hat{t}_i^+)\}_{i=1}^{N}$. Then by adjusting the number of output neurons, the ELM can directly generate the lower and upper bounds for a certain PINC. A structure of an ELM model for interval prediction is shown in FIGURE 1.

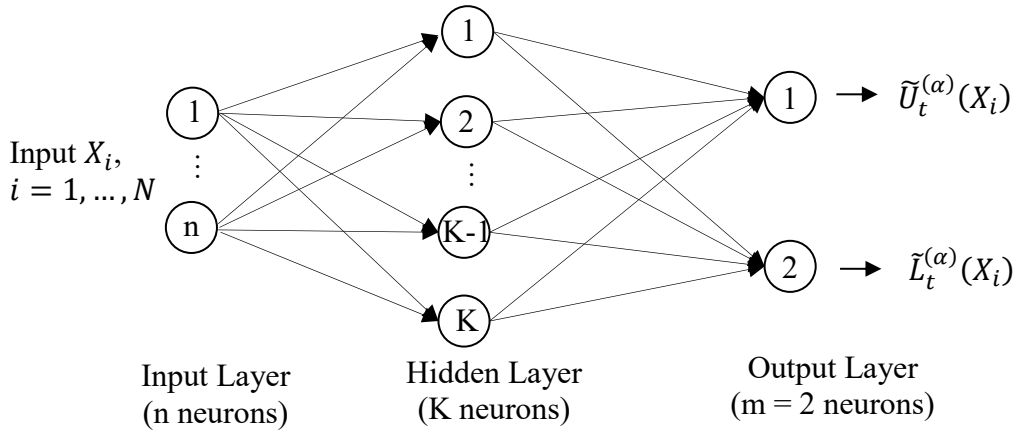

**FIGURE 1 A structure of ELM model for interval prediction.**

*ELM Improvement using Particle Swarm Optimization*
In this study, the Particle Swarm Optimizaiton (PSO) algorithm was used to further improve ELM, by minimizing a multi-objective optimization function which considers both reliabililty and sharpness of PIs. Specifically, a multi-objective optimization function was constructed to achieve the trade-off between those two important criteria. Recall that in ELM, the weights $\beta$ at the links connecting the hidden layer and output layer are the only parameters need be learned, which can be calculated in Equation (15). However, the output weights can be further tuned through Particle Swarm Optimization to minimize the following multi-objective function $F$.

$$\min_{\beta} F = \gamma A_t^{(\alpha)} + \lambda \bar{S}_t^{(\alpha)}{}_{norm} \tag{16}$$

Where, $A_t^{(\alpha)}$ is denotes the reliability as calculated by Equation (4),

$\bar{S}_t^{(\alpha)}{}_{norm}$ denotes sharpness as calculated by Equation (7).

$\gamma$ and $\lambda$ are trade-off weights for the reliability and sharpness metrics defined by the users. Some researchers have pointed out that reliability is the primary feature reflecting the correctness of the PIs, and hence should be given priority *(10)*.

Particle Swarm Optimization (PSO) is a population based heuristic optimization inspired by the social behavior of bird flocking or fish schooling *(16)*. It is an extremely simple but efficient algorithm with fast convergence speed for optimizing a wide range of functions. In this paper, it is applied to further adjust the output weights $\beta$ of ELM model in order to minimize the multi-object function in Equation (16). A brief introduction of PSO is given as following *(10)*.

Suppose the total population of particles in the $S$-dimensional search space is $N_P$, the position of the $i^{th}$ particle can be represented with a vector $P_i = [P_{i1}, P_{i2}, \ldots, P_{iS}]^T$. Once the algorithm starts learning, each particle is moving around in the space with a speed $v_i$. The algorithm keeps running until the user defined number of iterations $N_{iter}$ or a sufficiently good fitness has been reached (e.g., change of object values from two continuous runs is less than a user-defined threshold). For each iteration, the velocity and position of each particle are updated as following equations:

$$v_i = wv_i + c_1 R_1 \left( P_i^b - P_i \right) + c_2 R_2 (P_g^b - P_i) \tag{17}$$

$$P_i = P_i + \phi v_i \tag{18}$$

for $i = 1,2, \ldots, N_P$.

Where, $w$ is the inertia weight,

$c_1, c_2, \phi$ are user-defined constants,

$R_1$ and $R_2$ are random numbers within $[0, 1]$,

$P_i^b$ is the best position for the particle $i$ that generates the smallest objective function value from the previous iterations,

$P_g^b$ is the best position among particles in the global swarm that produces the smallest objective function value from the previous iterations.

Note that the velocity of the $i^{th}$ particle for the next iteration is a function of three components: the current velocity, the distance between its own previous best position $P_i^b$ and the current position, and the distance between the global best position $P_g^b$ and its current position. The initialized positions of the particles are generated randomly, based on the output layer weights $\beta^*$ using Equation (15), and the speed of the particles are randomly produced with an interval $[-v_{max}, v_{max}]$, $v_{max}$ is a $S$-dimensional vector. For each iteration, the updated position of each particle will be taken as the adjusted output layer weights $\beta$. The corresponding value from Equation (16) will be used to decide the $P_i^b$ and $P_g^b$. After the algorithm stops, the $P_g^b$ will be the finalized output layer weights for ELM model. The flow chart in FIGURE 2 shows the complete learning process of the hybrid PSO-ELM algorithm for interval prediction.

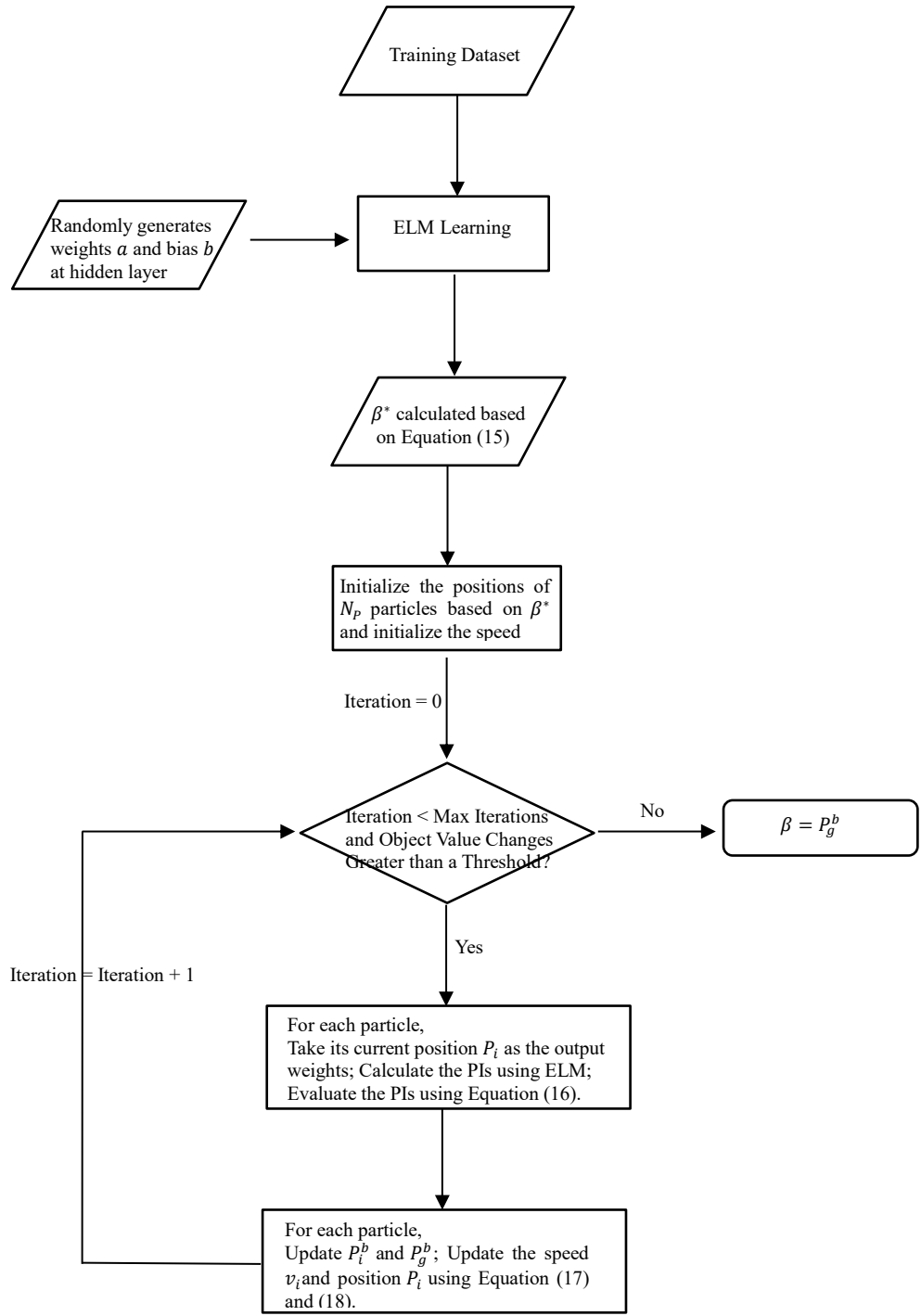

**FIGURE 2  The flow chart of PSO-ELM algorithm for interval prediction.**

**Benching marking PSO-ELM against ARMA Model and Kalman Filter**
To assess the performance of the proposed PSO-ELM model, it was compared against both an ARMA and a Kalman Filter model. ARMA has been applied for point prediction of short-term traffic volume for years *(2, 7)*. Given space limitations, the reader is referred to reference *(17)* for more details about the model. The Kalman Filter is also a very popular approach for short-term traffic volume prediction, having being first applied by Okutani and Yorgos in 1980s *(18)*. Since then, with its capability of conducting on-line learning and calibration, Kalman Filter based approaches have been applied and tested by quite a few traffic volume prediction studies *(19-20)*. However, again few studies have focused on the performances of interval prediction using Kalman filter. The reader is referred to references *(20)* and *(21)* for more details about Kalman Filter formulation and use for quantifying uncertainty through the generation of prediction intervals.

## DATASET DESCRIPTION

The authors of this paper have long studied the problem of predicting border crossing delay at the Peace Bridge *(22-26)*. Our previous work has clearly demonstrated that the traditional, single-value prediction approach cannot capture the dynamics of border crossing traffic volumes very well *(23-24)*. Instead a prediction interval approach, combining reliability and sharpness may be more appropriate. In the current study, we use a short-term traffic volume dataset comprised of hourly passenger car traffic volumes collected at the Peace Bridge focusing on traffic entering the US from Canada. The size of the dataset is 900 observations, collected between 7:00 to 21:00 from January 1st to March 1st in 2014. The first 600 data points (01/01/2014-02/09/2014) are used to train the models (i.e., the training dataset), while the rest (02/10/2014-03/01/2014) are used to test the models. Note that in this study, our object is to test and compare the interval prediction performances for different models, a smaller dataset is much easier for us to explore the reasons behind the outliers such as sport games, weather and so on, which will be discussed in detail in results analysis section.

## MODEL DEVELOPMENT AND RESULTS

### Model Development

The PSO-ELM model was implemented in Matlab. Multiple experiments were conducted to determine the parameters of the PSO-ELM model. First, target values in the training dataset were slightly increased and decreased by 5% in order to construct the target bounds $\{(\hat{t}_i^-, \hat{t}_i^+)\}_{i=1}^N$ (based on our experiments, this value doesn't have too much impact on the results). Then, for different PSO-ELM models built for three PINC levels (90%, 95% and 99%), we found they share the same configuration. The best ELM identified consisted of 14 neurons for the input layer, 20 neurons for the hidden layer, and 2 the for output layer. For the PSO algorithm, the population number $N_P$ was set to 50, the iteration times $N_{iter}$ to 150, and $w$, $c_1$ and $c_2$ in Equation (17) were set as 0.9, 1 and 1, respectively. The optimal value for $\phi$ in Equation (18) was 0.5, and the maximum particle speed $v_{max}$ was 2. However, as mentioned earlier, for the weights $w_1$ and $w_2$ in the sharpness calculation Equation (6), the optimial values are tuned carefully for different PINC levels, when the PINC was set to 90%, 6 and 0.1 were used, when the PINC was 95%, 11 and 0.1 were used, and when PINC was 99%, 12 and 0.1 were chosen. In general, we found larger $w_1$ will generate a narrower interval, and larger $w_2$ will make the intervals wider. Because $\alpha$ in Equation (6) decreased from 0.10 to 0.01 when PINC changed from 90% to 99%, we need to increase $w_1$ in order to keep the predicted interval tight. Finally, for the multi-objective function Equation (16), the weights of reliability and sharpness $\gamma$ and $\lambda$ were both set to 1 for all three PINC levels, which means that different with the wind power generation interval prediction *(10)* in our study, both criteria are regarded as equally important.

FIGURE 3 shows the values of the objective function, and the reliability and sharpness metrics as a function of the number of iterations.   As can be seen, the objective function value decreased from 0.79 to 0.21, the absolute average coverage error (AACE), the measure of reliability dropped from 0.506 to 0.037 with a clear declining trend (recall lower values of AACE indicated higher reliability or accuracy), and the sharpness curve fluctuated up and down but stabilized at around 0.17 level finally. The changes of the curves show that PSO can improve ELM to minimize the multi-objective function value.

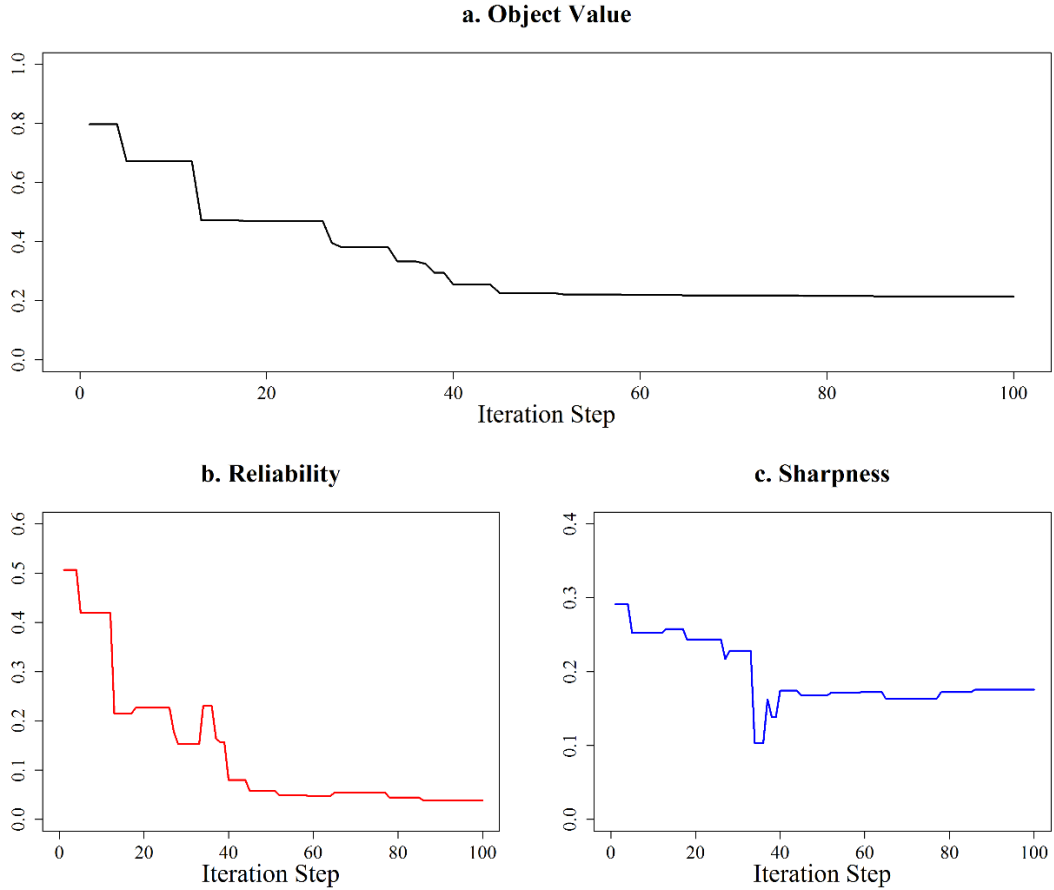

**FIGURE 3  Optimization curves in PSO-ELM algorithm with 95% PINC (a. change of object value; b. change of reliability; c. change of sharpness).**

For ARMA, the R package *forecast* was utilized to build the model *(27)* and the package was used to automatically select the optimal ARMA model for the given time series. The ARMA model's paramters were estimated used the training set of the first 600 observations, the parameters were then fized, and predictions were made for the next 300 points constituting the testing dataset.  For each data point or step, the prediction intervals with different PINCs 90%, 95% and 99% were calculated. For the Kalman Filter, on the other hand, all modeling and fitting were done using the 'R' package *dlm (21)*. The recursive form of the computations makes Kalman Filter natural and straightforward to compute the one-step-ahead forecasts and to update them sequentially as new data become available, e.g. the hourly traffic volumes arrive sequentially in short-term traffic prediction. Therefore, with the convenient form for on-line real time learning, it is of interest in this study to compare Kalman Filter to our off-line PSO-ELM model and ARMA model.

**Model Results**

After getting the PIs using PSO-ELM, ARMA and Kalman Filter models for different PINC levels, the reliability and sharpness can be calculated using Equation (4) and (7), as well as the minimum object value using Equation (16). TABLE 1 summarizes the performance of the three models considered namely, the proposed PSO-ELM, ARMA and Kalman Filter. The table shows the values of the reliability and sharpness metrics, the value of the multi-objective function. It also lists the actual PI Coverage Probability (PICP), which is the ratio of the 300 observations in the testing dataset falling within the PIs, and the mean PI length (MPIL) which is the average distance between the upper bound and lower bound of the intervals *(13)*. These can help us to evaluate the models for the further step in a more straitforward way.

**TABLE 1 Model Performances under Different PINC Levels**

| PINC | Models | Reliability | Sharpness | Object Value | PICP | MPIL |
|------|--------|-------------|-----------|--------------|------|------|
| 90% | PSO-ELM | 0.003 | 0.180 | 0.183 | 90.33% | 255.96 |
| | ARMA | 0.060 | 0.264 | 0.324 | 84% | 374.71 |
| | Kalman Filter | 0.030 | 0.142 | 0.172 | 87% | 204.70 |
| 95% | PSO-ELM | 0 | 0.190 | 0.190 | 95% | 295.33 |
| | ARMA | 0.060 | 0.288 | 0.348 | 89% | 446.50 |
| | Kalman Filter | 0.020 | 0.154 | 0.174 | 93% | 243.90 |
| 99% | PSO-ELM | 0.003 | 0.056 | 0.059 | 99.33% | 396.90 |
| | ARMA | 0.040 | 0.076 | 0.116 | 95% | 586.80 |
| | Kalman Filter | 0.017 | 0.043 | 0.060 | 97.33% | 320.43 |

The results in TABLE 1 reveal the following interesting observations. First, from the aspect of reliability, PSO-ELM has the smallest error values for all three PINC levels (and thus appears to be the most reliable), followed by Kalman Filter and then ARMA. Except that, PSO-ELM model is the only one the PICPs of which are actually higher than or equal to the corresponding PINCs. Second, not surprisingly, the sharpness has a positive correlation with the MPIL. A lower sharpness value means a smaller MPIL. In this comparison, on average Kalman Filter produces a little narrower prediction intervals than PSO-ELM (a difference of 50 to 70 vehicles/hour based on MPIL). ARMA once again performs the worst with an obvious gap. Third, the column of objective value which is the summation of reliability and sharpness shows that for 90% and 95% PINC levels, the best performing model is the Kalman Filter model, closely followed by PSO-ELM model. For 99% PINC level, the objective function values of PSO-ELM and Kalman Filter are almost the same (0.059 vs. 0.060). ARMA has a much higher objective value. Last but not the least, considering that the PSO-ELM model is applied an off-line fashion (trained based on the 600 observations, and the parameters are kept unchanged for the next 300-step predictions) whereas the Kalman Filter is used to make one-step-ahead predictions and updated every step, it seems that the results for the proposed PSO-ELM are very good. In fact, for reliability, PSO-ELM appears to be the best model.

FIGURE 4 visualizes the prediction intervals of PSO-ELM under the three different confidence levels for the 300-data point test set. As can be seen, going from the top figure to the one at the bottom, when the PINC level increases from 90% to 99%, the prediction intervals become correspondingly wider and more observations fall within the prediction intervals. For example, the point marked with black circle is outside of the prediction interval under 90% PINC, but within the PIs under 95% and 99% levels. Similarly, the point marked with orange circle is only covered by the prediction interval when the PINC is 99%.

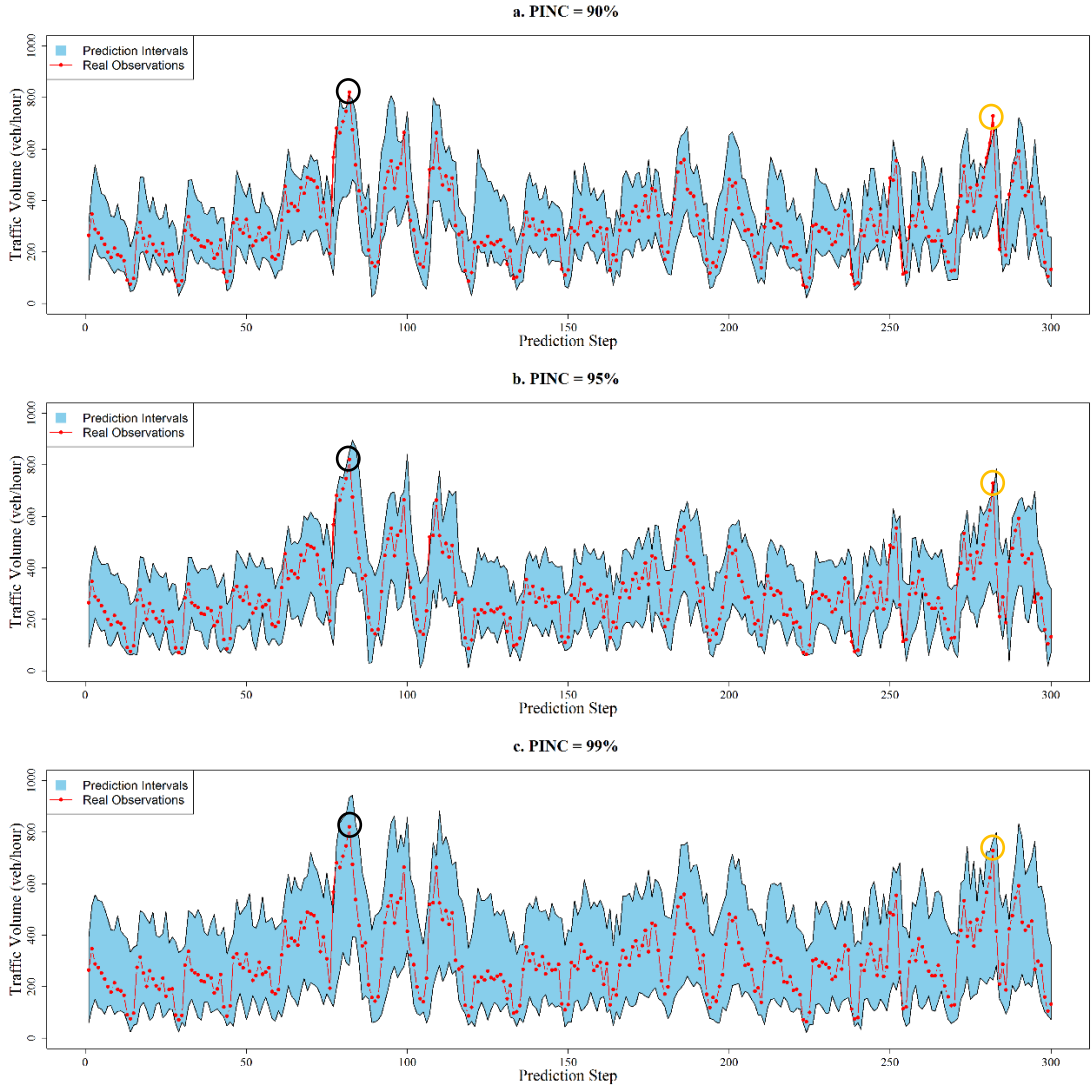

**FIGURE 4  PIs of PSO-ELM by PINC levels (a. PINC = 90%; b. PINC = 95%; c. PINC = 99%).**

FIGURE 4 also shows that even for the points that fell outside of the PIs given by PSO-ELMs, the distances between those points and the interval bounds are very small. In order to verify this, TABLE 2 lists the average distances between the points lying outside the PI and the upper or lower bounds for the three PINC levels.

**TABLE 2 Average Distance to Bound for Points outside of PSO-ELM PIs**

| PINC | Relationship with PI | Number of Points | Average Distance to Bound |
|---|---|---|---|
| 90% | Greater than Upper Bound | 16 | 39.23 |
| | Smaller than Lower Bound | 13 | 26.65 |
| 95% | Greater than Upper Bound | 7 | 52.63 |
| | Smaller than Lower Bound | 8 | 25.38 |
| 99% | Greater than Upper Bound | 1 | 80.18 |

| | Smaller than Lower Bound | 1 | 5.93 |
|---|---|---|---|

As shown in TABLE 2, the outlying points are still very close to the upper bounds or the lower bounds. For example, when PINC is 90%, there are 29 points outside of the PIs in total. For the 16 points greater than the upper bounds, the average distance is about 39 vehicles/hour, For the 13 points smaller than the lower bounds, the average distance is only 26 vehicles/hour.

The points lying outside the bounds were closely examined to discern possible reasons that led to this happenning. As an example, when the PINC was 90%, we checked each of the 29 points lying outside of the PIs from the aspects of the day of the week, peak hour, and weather and whether the day had a special event (e.g., a sport games occurred on that day). For the day of the week, we distinguish between weekends (long weekend holiday, Friday, Saturday and Sunday) and non-weekends. Peak hours are defined as those between 07:00-09:00 and 17:00-19:00 for non-weekends. For weather, we distinguish between snowy and non-snowy days based on precipitation information downloaded through the Weather Underground website *(28)*. For special event days, we consider whether a major sport game was taking place that day (included aer home games of the Buffalo Bills and Buffalo Sabres). FIGURE 5 shows the situations under which the 29 outliers occurred.

### a. 16 Points Greater than the Upper Bound

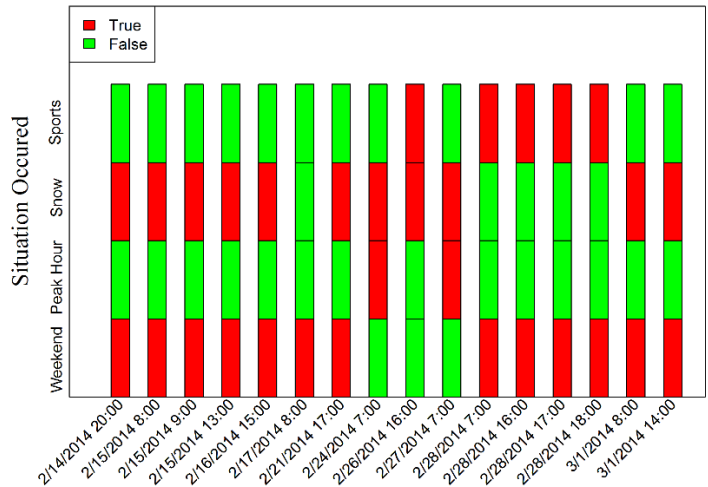

### b. 13 Points Smaller than the Lower Bound

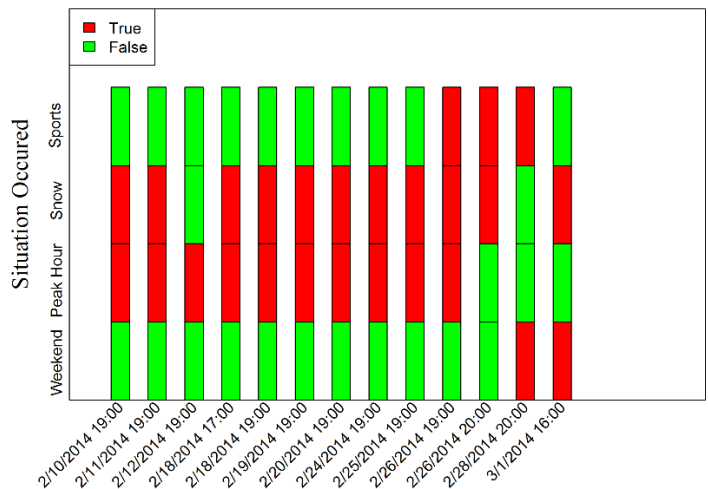

**FIGURE 5 Exploration of 29 Points Outside of PSO-ELM PIs When PINC = 90% (a. 16 Points Greater than Upper Bound; b. 13 Points Smaller than Lower Bound)**

First, from FIGURE 5 we can see that for the data set considered, it had snowed except three days (Feb 12th, 17th, and 28th), and therefore it is hard to judge whether snow has an impact or not. Second, from FIGURE 5a, the majority of the points that were beyond the upper bound of the interval (specifically 13 out of the 16 points) occurred at weekends and Presidents' Day (Feb 17th). With respect to FIGURE 5b for the 13 points that were below the lower bound, it can be seen that the majority occurred on weekdays (10 out of the 13 points), and during the peak hours in the afternoon. Third, regarding special events, there were two Buffalo Sabres matches on Feb 26th and Feb 28th in 2014. Based on the official website *(29)*, the game schedule was at 19:30 and 19:00, respectively. Because Canadian fans need to travel to the US before the game starts, the traffic volume at 16:00 on Feb 26th, for example, was greater than the upper bound of PI (FIGURE 5a), whereas the traffic volumes during the game time itself (e.g., at 19:00 and 20:00) on the same day were smaller than the lower bound (FIGURE 5b). Similar traffic patterns could be observed because of the other game on Feb 28th.

**CONCLUSIONS AND FUTURE RESEARCH DIRECTIONS**
In this study, we introduced and applied a hybrid machine learning model called PSO-ELM for interval prediction of short-term traffic volume. The dataset considered was an hourly traffic data set for traffic crossing the Peace Bridge International Border. The paper compared the performance of the PSO-ELM model against two other probabilistic-based approaches ARMA and Kalman Filter, focusing on reliability and sharpness metrics for the PIs. The main findings are summarized as following:

1. The PSO-ELM model requires no statistical inference or distribution assumption of the error item, unlike the probabilistic based interval forecasting methods ARMA and Kalman Filter. The PSO-ELM algorithm showed a superior performance in terms of minimizing the value of the multi-objective function evaluating both reliability and sharpness, thanks to the tuning of the weights of the output layer of ELM by the PSO algorithm.

2. By comparing the objective values of the multi-objective function for the three models under different PINC levels, ARMA models appeared to perform far behind PSO-ELMs and Kalman Filter. For the 90% and 95% PINC levels, the Kalman Filter model was closely followed by PSO-ELM model. For 99% PINC level, PSO-ELM was a little better than the Kalman Filter. The Kalman Filtering model had a slightly smaller MPILs than PSO-ELM models. However, only the PICPs of PSO-ELM models were actually higher than or equal to the corresponding PINCs. Considering that PSO-ELM model was implemented in an off-line fashion, whereas the Kalman Filter model is an on-line model, the results of the proposed PSO-ELM appears quite promising.

3. The visualizations of the PIs given by PSO-ELM show that, as expected, the width of the PIs increases when the PINC level increases from 90% to 99%. The average distance to the bounds for the few points lying outside of PSO-ELM PIs are verified to be quite small.

4. Outliers that are higher than the upper bound of the PI appear to occur on weekends and holidays, whereas outliers lying below the lower tend to occur on weekdays and during the peak hour in the afternoon.

5. The schedules of sports games can lead to traffic levels before the games higher than upper bounds of the PI, and traffic levels during the game time smaller than the lower bounds.

For future research, the researchers plan to investigate the possibility of developing an on-line version of the PSO-ELM model. We also plan to include additional variables to capture the effect of inclement weather and special events (those could be discovered from mining social media data) to improve the prediction models accuracy *(28, 30)*. It also may be of interest to identify the most frequent or recurring traffic patterns at the border using association rules learning *(31)* and frequent pattern tree learning *(32-33)*.